%% file: main.tex
\documentclass[10pt,twocolumn,letterpaper]{article}

\usepackage[pagenumbers]{cvpr} 

\definecolor{cblue}{rgb}{0.21,0.49,0.74}
\usepackage[pagebackref,breaklinks,colorlinks,allcolors=cblue]{hyperref}
\usepackage[sort&compress]{natbib}
\usepackage{tabularx}
\usepackage{multirow}
\usepackage{textcomp}
\usepackage{adjustbox}

\title{ATP-LLaVA: Adaptive Token Pruning for Large Vision Language Models}

\author{Xubing Ye\textsuperscript{1},
        Yukang Gan\textsuperscript{2}\textsuperscript{$\dagger$}, 
        Yixiao Ge\textsuperscript{2*}, 
        Xiao-Ping Zhang\textsuperscript{1}, 
        Yansong Tang\textsuperscript{1*} \\
$^{1}$Tsinghua Shenzhen International Graduate School, Tsinghua University, $^{2}$ARC Lab, Tencent PCG\\
{\tt\small \{yxb23@mails.,tang.yansong@sz.,xiaoping.zhang@sz.\}tsinghua.edu.cn} \\
{\tt\small \{brucegan,yixiaoge\}@tencent.com} 
}

\begin{document}
\maketitle

\renewcommand{\thefootnote}{}
\footnote{Work was done when the author interned at ARC Lab, Tencent PCG. $^{\dagger}$Project lead. $^{*}$Corresponding author.}
\renewcommand{\thefootnote}{\arabic{footnote}} 

\input{sec/0_abstract}

\input{sec/1_intro_related}

\input{sec/2_method}

\input{sec/3_experiment}

\input{sec/4_conclusion}
{
    \small
    \bibliographystyle{ieeenat_fullname}
    \bibliography{main}
}


\end{document}

%% file: sec/0_abstract.tex
\begin{abstract}
Large Vision Language Models (LVLMs) have achieved significant success across multi-modal tasks. 
However, the computational cost of processing long visual tokens can be prohibitively expensive on resource-limited devices. 
Previous methods have identified redundancy in visual tokens within the Large Language Model (LLM) decoder layers and have mitigated this by pruning tokens using a pre-defined or fixed ratio, thereby reducing computational overhead.
Nonetheless, we observe that the impact of pruning ratio varies across different LLM layers and instances (image-prompt pairs).
Therefore, it is essential to develop a layer-wise and instance-wise vision token pruning strategy to balance computational cost and model performance effectively.
We propose ATP-LLaVA, a novel approach that adaptively determines instance-specific token pruning ratios for each LLM layer.
Specifically, we introduce an Adaptive Token Pruning (ATP) module, which computes the importance score and pruning threshold based on input instance adaptively.
The ATP module can be seamlessly integrated between any two LLM layers with negligible computational overhead.
Additionally, we develop a Spatial Augmented Pruning (SAP) strategy that prunes visual tokens with both token redundancy and spatial modeling perspectives.
Our approach reduces the average token count by 75\% while maintaining performance, with only a minimal 1.9\% degradation across seven widely used benchmarks.
The project page can be accessed via the following \href{https://yxxxb.github.io/ATP-LLaVA-page/}{link}.
\end{abstract}

%% file: sec/1_intro_related.tex
\section{Introduction}
\label{sec:intro}

\begin{figure}[t]
\centering
\includegraphics[width=1\columnwidth]{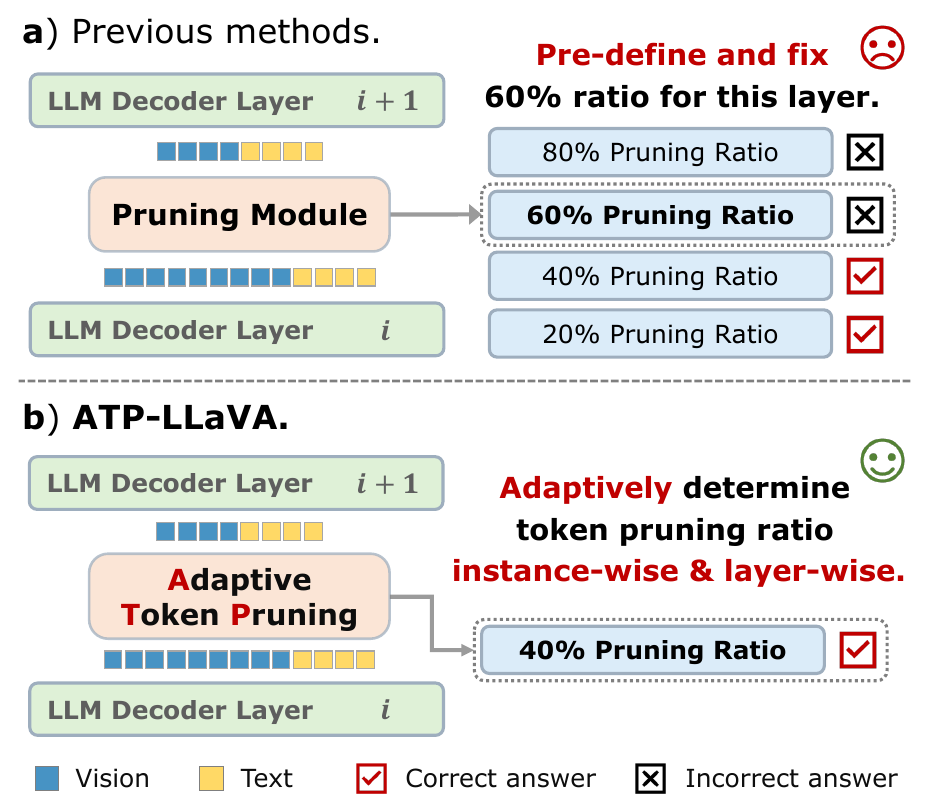}
\vspace{-14pt}
\caption{
(a) Previous methods employ a fixed, pre-defined token pruning ratio.
(b) Illustration of ATP-LLaVA, which dynamically selects the adaptive pruning ratio for each layer of the LLM decoder based on the instance-specific characteristics.
}
\vspace{-8pt}
\label{fig:i1-1}
\end{figure}

The emergence of Large Vision Language Models (LVLMs)~\cite{zhu2023minigpt, li2023blip2, liu2023llava, du2022glm, Qwen-VL, internlmxcomposer, liu2023improvedllava, fuyu2023, liu2023world, geminiteam2024gemini} has significantly advanced visual understanding. 
These approaches leverage visual encoders to extract visual features, which are then processed jointly with text in Large Language Models (LLMs). 
Despite their impressive multi-modal understanding capabilities, the deployment of these models is often hindered by the substantial memory and computational costs when processing large number of visual tokens, especially in resource-constrained environments.

To address this issue, previous methods~\cite{cha2023honeybee, shang2024LLaVA-PruMerge, ye2024voco, chen2024image} have focused on compressing visual tokens by pruning redundant ones, as visual information tends to be sparser compared to natural language information. 
While these methods can mitigate the loss of model's understanding capabilities caused by token pruning, they share a common limitation: requiring a fixed pruning ratio (\emph{i.e.}, a fixed number of retained tokens or a non-learnable threshold) to be \textbf{pre-defined} for the model, as shown in~\cref{fig:i1-1} (a).
Recently, some methods~\cite{hu2024matryoshka, cai2024matryoshka} have explored training a single model that can handle varying visual token counts.
However, it remains necessary to manually specify a pruning ratio for each individual sample.
Due to the varying complexity of different tasks and the differing levels of content within images, a pre-defined pruning ratio may result in either information loss or excessive information retention, consequently impacting the model's efficiency and effectiveness.

\begin{figure}[t]
\centering
\includegraphics[width=1\columnwidth]{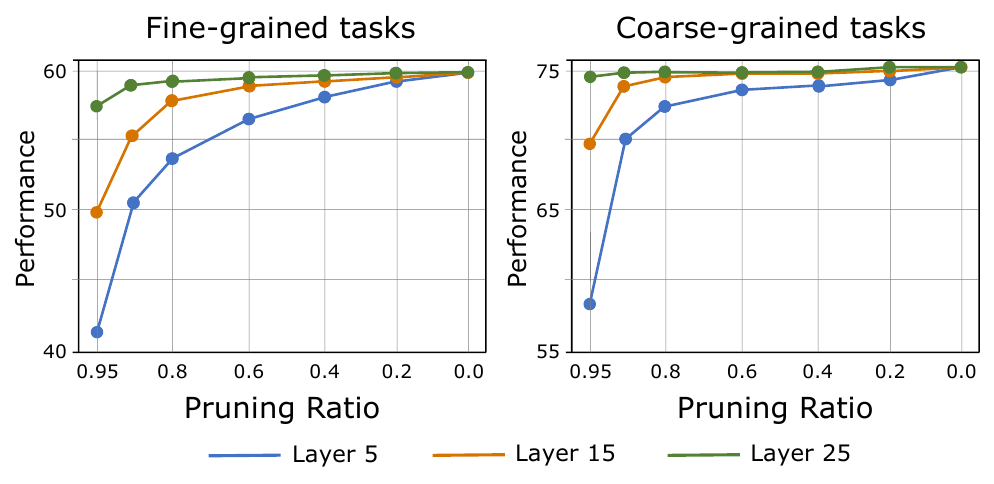}
\vspace{-12pt}
\caption{
Comparison of vision token pruning at different LLM decoder layers and pruning ratios across fine-grained and coarse-grained tasks of the SEED-Image~\cite{li2023seed}.
Fine-grained tasks include instance counting, spatial relation, etc. Coarse-grained tasks include scene understanding, etc.
}
\vspace{-8pt}
\label{fig:i1-2}
\end{figure}

To further validate this, we conducted a preliminary experiment to examine the effects of pruning visual tokens at various layers of LVLMs and across different pruning ratios. As shown in~~\cref{fig:i1-2}, the impact of pruning ratio on performance is \textbf{\textit{layer-wise}}, with shallower layers being more sensitive to pruning and deeper layers exhibiting greater robustness.
Moreover, we compared the performance of pruning across tasks of varying complexity and observed that the pruning ratio's impact is also \textbf{\textit{instance-wise}}.
Fine-grained tasks, such as instance counting and spatial relation, demand detailed visual information, necessitating the retention of more tokens at each layer to prevent performance degradation. 
In contrast, coarse-grained tasks like scene understanding do not exhibit significant performance loss even at high pruning ratios. 
These observations indicate that the optimal pruning ratio, which achieves the best balance between performance and efficiency, varies for each instance and each layer. 
A pre-defined pruning ratio can lead to suboptimal model performance and efficiency.

In this paper, we propose a framework called \textbf{A}daptive \textbf{T}oken \textbf{P}runing for large vision language models (\textbf{ATP-LLaVA}), as illustrated in~\cref{fig:i1-1} (b), which adaptively determines the pruning ratio. Specifically, 
we design an Adaptive Token Pruning (ATP) module that can be seamlessly integrated between any two LLM layers with negligible computation cost.
The ATP module first employs a Spatial Augmented Pruning (SAP) policy to dynamically prune tokens for each instance. This policy adopts scores on two perspectives to evaluate the importance of each token.
The first perspective, the redundant pruning score, computes token importance by leveraging both intra-modal and cross-modal correlations. 
The second perspective, the spatial pruning score, evaluates the spatial information within the token sets that are spatially uniform sampled at various granularity.
Building upon these two scores, two learnable thresholds are introduced at each ATP module to dynamically select important tokens for each LLM layer and each instance. 
Tokens that are not selected will be discarded in subsequent layers.
Lastly, to further enhance the training process, we propose an ATP-Loss function to strike a balance between token pruning efficiency and understanding capability. 
We summarize our contributions as follows:
\begin{itemize}
\item 
We reveal the importance of adaptively determining pruning ratios at the instance and LLM layer levels for effective visual token pruning, and propose ATP-LLaVA, a framework that dynamically reduces computational cost for large vision language models.
\item 
To enable the model to learn the adaptive token pruning strategy, we introduce an Adaptive Token Pruning (ATP) module, along with an ATP-Loss function, thereby balancing pruning efficiency and model capabilities.
\item 
To mitigate the loss in visual understanding performance caused by token pruning, we introduce a Spatial Augmented Pruning (SAP) approach
, which preserves spatial modeling during the pruning process.
\item 
ATP-LLaVA achieves a 75\% average pruning ratio while maintaining 98.1\% performance across seven widely used vision understanding benchmarks.

\end{itemize}

\section{Related work}
\textbf{Large Vision-Language Models.}
The impressive progress of large language models (LLMs)~\cite{vicuna2023, gpt3, gpt4, workshop2023bloom, touvron2023llama, Hugo2023, jiang2023mistral, huang2024good} has sparked interest in developing large vision language models (LVLMs) that can bridge the gap between visual and linguistic understanding. LVLMs have shown impressive capabilities in cross-modal understanding and visual language tasks through modality alignment and instruction tuning.
Previous works~\cite{li2023blip2, liu2023llava, zhu2023minigpt, du2022glm, Qwen-VL, internlmxcomposer, yang2024language, bai2024self, liu2023improvedllava, fuyu2023, instructblip, alayrac2022flamingo, Qwen2VL} have validated the efficacy of this training paradigm in visual understanding.
The success of LVLMs has also been extended to the video domain~\cite{liu2023world, geminiteam2024gemini, li2023llamavid, huang2023vtimellm, jin2024chatunivi, liu2022learning, liu2022global, luo2024soc, damonlpsg2023videollama, 2023videochat, liu2023btadapter, luo2023valley, Maaz2023VideoChatGPT, zhang2024flashvstreammemorybasedrealtimeunderstanding}.
Furthermore, research has shown that LVLMs can capture rich visual information for understanding and generation when provided with high-resolution images~\cite{liu2023improvedllava, fuyu2023}.
However, the growing number of vision tokens occupy a substantial portion of the LLM's valuable context window and leading huge bottleneck for computational infrastructure.
To address this, further innovation in token compression and pruning techniques is essential.

\noindent\textbf{Vision Token Compression and Pruning.}
To compress vision information with less tokens, previous methods~\cite{li2023blip2, zhu2023minigpt, du2022glm, internlmxcomposer, instructblip} have largely employed Q-Former~\cite{li2023blip2}, which maps images to fixed-length learnable queries. \cite{li2023llamavid, yao2024decodecouplingtokencompression} have applied simple pooling strategy to downsample visual features.
\cite{ye2024voco} try to distill the LLMs' understanding paradigm of vision tokens into single VoCo token to reduce inference cost.
\cite{shang2024LLaVA-PruMerge} identifies redundant visual tokens through clustering analysis and prunes them.
\cite{chen2024image} reveals that vision tokens within LLM Transformer layers are also redundant, and pruning them internally incurs less penalty than pruning before inputting to the LLM.
Although these methods can mitigate the loss caused by token compression, they are limited by relying on a pre-defined pruning rate.
While~\cite{rao2022dynamicvit, kim2022learned, cao2024madtp} explore adaptive token pruning in Vision Transformers, this area remains relatively under-explored in the context of Large Vision Language Models.
Recent methods~\cite{cai2024matryoshka, hu2024matryoshka} offer flexible choices for the number of visual tokens, but they struggle to adaptively determine the optimal pruning ratio.
ATP-LLaVA can adaptively determine the pruning ratio within any layer of the LLM, based on the specific instance characteristics.

%% file: sec/2_method.tex
\section{Method}
\label{sec:method}

\subsection{Overview}

The ATP-LLaVA architecture, as shown in~\cref{fig:i2}, centers around the \textbf{A}daptive \textbf{T}oken \textbf{P}runing (ATP) module that can be easily inserted between any two decoder layers in the LLM decoder backbone. 
The ATP module first computes importance scores for visual tokens based on the self-attention map from the previous layer, and then utilizes two lightweight prediction heads to learn pruning thresholds with minimal additional parameters.
These two prediction heads generate learnable thresholds for redundant and spatial pruning of visual tokens, respectively, enabling layer-wise and instance-wise adaptive pruning.

In the following sections, we provide a brief overview of the LLM decoder in~\cref{sec:prelimi}.
We then present our Adaptive Token Pruning module in~\cref{sec:pro}. 
Specifically, the ATP module consists of three key components: Spatial Augmented Pruning approach (\cref{sec:pruning}), Learnable Thresholds (\cref{sec:learnable}), and ATP-Loss (\cref{sec:loss}).

\subsection{Preliminaries}
\label{sec:prelimi}
The LLM decoder consists of multiple decoding layers (\emph{i.e.}, 32 layers in LLaMA~\cite{touvron2023llama}). 
Each decoding layer typically uses self-attention with a causal mask.
The input to each Transformer layer is a concatenation of vision and text tokens $H\in \mathbb{R}^{L \times D}$, where $L$ is the total length of the tokens and $D$ is the hidden dimension.
For the causal attention mechanism (considering single-head attention as an example), the self-attention logits can be computed by 
\begin{align} 
\label{eq:1}
{A}_{\text{logits}} = {Q} {K}^T / \sqrt{D},
\end{align}
where ${Q} \in \mathbb{R}^{L \times D}$ and ${K} \in \mathbb{R}^{L \times D}$ are the query and key matrices, respectively. 
It yields the resulting logits matrix ${A}_{\text{logits}} \in \mathbb{R}^{L \times L}$.
To accommodate the causal attention mechanism in the decoder, a lower triangular masking matrix $M \in \mathbb{R}^{L \times L}$ is element-wise added to the logits when computing final attention weights ${A}$, resulting in 
\begin{align} 
\label{eq:2}
{A} = \text{Softmax}({A}_{\text{logits}} + M),
\end{align}

Following the self-attention, each layer is connected to a Feed Forward Network (FFN) layer, enabling the model to capture contextual relationships within each modality.

\subsection{Adaptive Token Pruning}
\label{sec:pro}
Given the output of the $i$-th Transformer decoder layer, denoted as $H_i$. 
It consists of two components: the hidden states of visual tokens, represented as ${V_i} \in \mathbb{R}^{L_v \times D}$, and the hidden states of text tokens, represented as ${T_i} \in \mathbb{R}^{L_t \times D}$ (we ignore the system prompt input for simplicity).
The ATP module removes redundant tokens from ${V}$, resulting in unpruned visual tokens, denoted as ${V_i^p} \in \mathbb{R}^{L_{v}^p \times D}$, where $L_{v}^p < L_v$. 
The maintained visual tokens are then concatenated with the original text tokens ${T}$ and fed into the following LLM layer.
Visual tokens pruned at the current layer are irretrievable in subsequent layers.

\begin{figure*}[ht]
\centering
\includegraphics[width=0.99\linewidth]{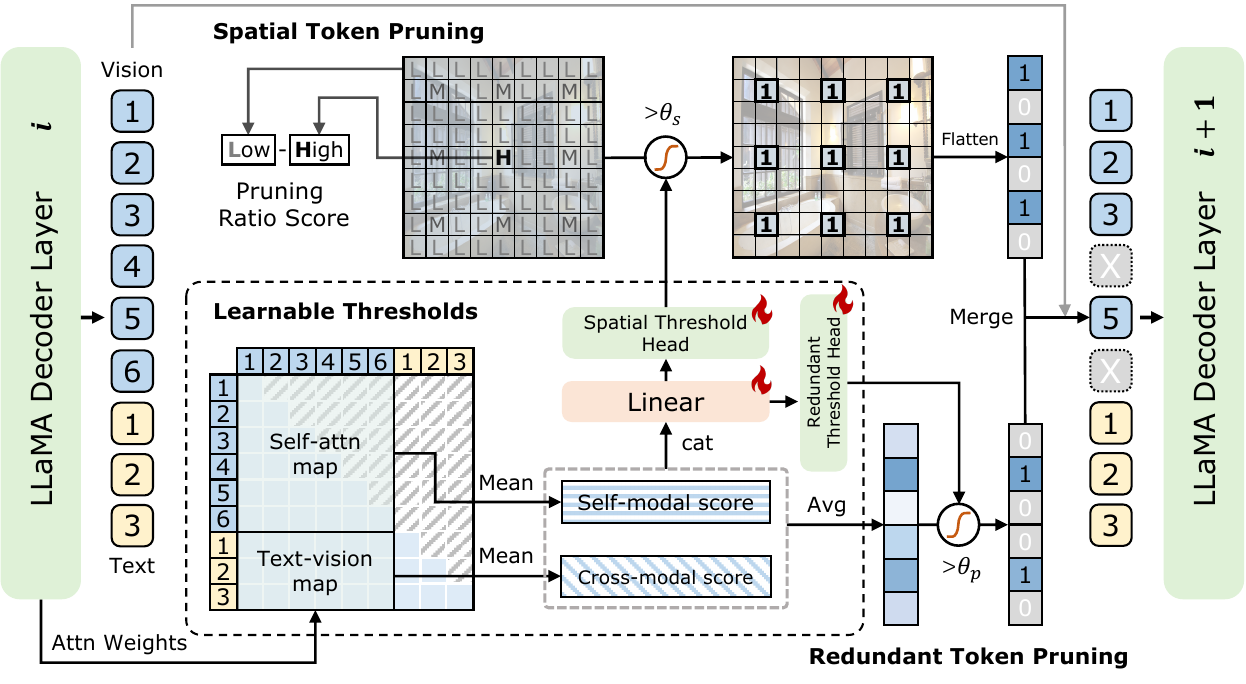}
\vspace{-6pt}
\caption{
Illustration of the \textbf{A}daptive \textbf{T}oken \textbf{P}runing (ATP) module. 
The ATP module can be flexibly inserted between any two LLaMA decoder layers.
It adaptively predicts pruning thresholds for current layer and instance.
Redundant or text-irrelevant visual tokens are pruned at this stage, and they will be ignored by other tokens in subsequent LLaMA decoder layers.
}
\vspace{-8pt}
\label{fig:i2}
\end{figure*}

\subsubsection{Spatial Augmented Pruning}
\label{sec:pruning}
Token pruning in large vision language models disrupts spatial modeling, which is crucial for vision understanding. 
To mitigate this, we introduce the \textbf{S}patial \textbf{A}ugmented \textbf{P}runing (SAP) approach.
SAP consists of two stages: (1) redundant pruning, which assigns importance scores to visual tokens and prunes them based on a threshold, and (2) spatial pruning, which uniformly samples tokens in the spatial dimension. 
The pruning masks from both stages are then combined to form the final pruning strategy.

\noindent\textbf{Redundant Pruning Score.}
To perform redundant token pruning, we define the importance score for each token. 
Intuitively, we consider both the vision token's self-modality importance and its importance to the text modality.
For the current layer's visual tokens ${V_i}=\{v_1, \ldots, v_{L_v}\}$, a visual token $v_n$ is deemed important if it receives significant attention from other visual tokens in the same layer.
Specifically, as shown in the \textit{self-attn map} of~\cref{fig:i2} (left), this importance score can be obtained from the attention logits ${A}_{\text{logits}}^i$ of the $i$-th Transformer layer. 
The self-modality importance score $S^{self}_n$ for token $v_n$ is defined as:
\begin{align} 
\label{eq:3}
S^{self}_n = \frac{1}{L_v} \sum_{m=1}^{L_v}{A}_{\text{logits}}^i (v_n, v_m) \in \mathbb{R}^{L_v} ,
\end{align}

Similarly, visual tokens that receive greater attention from all text tokens should be assigned a higher importance score in the text modality.
As shown in the \textit{text-vision map} of~\cref{fig:i2} (left), this importance score can be obtained from the attention map ${A}^i$ of the $i$-th Transformer layer. 
Given the current layer's text tokens ${T_i}=\{t_1, \ldots, t_{L_t}\}$, we define the cross-modality importance score $S^{cross}_n$ for token $v_n$ as:
\begin{align} 
\label{eq:4}
S^{cross}_n = \frac{1}{L_t} \sum_{m=1}^{L_t}{A}^i (v_n, t_m)  \in \mathbb{R}^{L_v} ,
\end{align}
where the probability ${A}^i (v_n, t_m)$ represents the normalized value that token $t_m$ focuses on $v_n$. 
We calculate the final score $S^{redundant}_n$ by taking the average of $S^{self}_n$ and $S^{cross}_n$.

Having obtained the token scores, a learnable pruning threshold is introduced to selectively retain tokens.
A detailed discussion on this is presented in~\cref{sec:learnable}.

\noindent\textbf{Spatial Pruning Score.}
Uniform spatial sampling of visual tokens was proposed by LLaVA-PruMerge~\cite{shang2024LLaVA-PruMerge} to preserve spatial visual information.
However, we observe that the impact of sampling ratio varies across different input instances and LLM layers. 
We found that excessive sampling of spatial tokens significantly increases token count, with the sampling ratio's impact varying across different input instances and LLM layers.
To address this, we design a dynamic uniform spatial sampling strategy.

As illustrated in~\cref{fig:i2} (top), we define a spatial pruning score $S^{spatial}\in \mathbb{R}^{L_v}$ within the range of (0, 1]. 
The sampling rate $R^{s}$ is defined as the ratio of the uniform sampled tokens to the visual tokens $L_v$. 
Under this sampling rate, we acquire a set of sampled visual tokens $V_{s}$. 
The score of the sampled visual tokens is defined as:
\begin{align} 
\label{eq:5}
S^{spatial}_{n} = 1 - R^{s} \cdot \lambda_{sample} ,\quad v_n \in V_s ,
\end{align}
where $\lambda_{sample}$ is a scaling coefficient. 
Tokens sampled at higher rates are given higher scores.
We introduce a learnable threshold for spatial pruning, allowing us to dynamically adjust the spatial pruning rate instance-wise and layer-wise, despite the fixed spatial pruning score.

\noindent\textbf{Positional Embeddings for Token Pruning.}
Merely applying uniform spatial sampling is insufficient to enhance spatial modeling, as the sampled tokens will be flattened into a sequence and fed into the LLM.
To address this, we employ 2D rotary embedding~\cite{su2023roformerenhancedtransformerrotary, Qwen2VL}, which enhances the spatial information.
Furthermore, previous methods typically reorganize the retained tokens' position embeddings into a contiguous sequence (\emph{e.g.}, $0, 1, 2, ...$), which may disrupt spatial features under 2D positional encoding. 
In contrast, we preserve the original position embeddings of the retained tokens after pruning.

\subsubsection{Adaptive Pruning with Learnable Thresholds}
\label{sec:learnable}
To adapt to the dynamic changes in pruning thresholds across different layers and instances, we introduce a MLP with dual prediction heads to learn instance-specific thresholds. 
The determination of pruning thresholds should be highly correlated with the instance itself, and thus we leverage the self-modality and cross-modality scores computed in~\cref{sec:pruning} as inputs to the prediction module.

Specifically, given the visual token scores $S^{self}_{n}$ and $S^{cross}_{n}$, we compute the redundant pruning threshold $\theta_{r}$ and spatial pruning threshold $\theta_{s}$ as follows:
\begin{align} 
\label{eq:6}
z &= \text{Linear}(cat(S^{self}_{V_i}, S^{cross}_{V_i})) , \\
\theta_{r} &= \sigma(\text{Linear}_r(z)) \in \mathbb{R}^{1} ,\\
\theta_{s} &= \sigma(\text{Linear}_s(z)) \in \mathbb{R}^{1} ,
\end{align}
However, the hard masks generated by threshold prevent gradient backpropagation, rendering the threshold prediction module untrainable.
Inspired by~\cite{bengio2013estimatingpropagatinggradientsstochastic}, we convert the hard masks to differentiable soft masks:
\begin{align} 
\label{eq:7}
Mask_{i}^r &= \sigma((S^{redundant}_{V_i} - \theta_{r}) \cdot T) \in \mathbb{R}^{L_v} ,\\
Mask_{i}^s &= \sigma((S^{spatial}_{V_i} - \theta_{s}) \cdot T) \in \mathbb{R}^{L_v} ,\\
Mask_{i} &= max (Mask_{i}^r, Mask_{i}^s) \in \mathbb{R}^{L_v} ,
\end{align}
where $T$ is a temperature coefficient that, when sufficiently large, renders the sigmoid function a differentiable mask.

Besides, directly discarding unselected visual tokens using the mask is non-differentiable for the learnable threshold module, hindering end-to-end learning of token pruning strategies. 
And pruning varying numbers of tokens per instance within a batch complicates parallel training.
To address these challenges, we followed~\cite{rao2022dynamicvit} and utilize masks during the attention \textit{Softmax} operation, effectively eliminating the influence of pruned tokens on others and ensuring a differentiable process.
Specifically, we multiply the exponential results by the pruning mask after computing the exponentials and before summing them up.
We only apply the softmax mask during the training phase. 
In the inference phase, the pruned tokens are directly discarded and do not participate in any subsequent layer computations.

\subsubsection{Budget-Constrained Training}
\label{sec:loss}
Our goal is to encourage the model to learn an optimal pruning strategy that balances computational cost and performance loss.
To achieve this, we design an ATP loss function that trains the model under limited pruning constraints.
Specifically, we introduce a penalty term that discourages the model from retaining excessive tokens.
The penalty term should be designed to increase with both the number of remaining tokens and the layer depth, accounting for the diminishing returns of deep pruning on computational cost.

Given the index set $I = \{i_0, \ldots, i_n\}$ of LLM layers where ATP module is introduced, the differentiable pruning masks are denoted as $Masks = \{Mask_{i_k} \mid i_k \in I\}$.
To facilitate backpropagation, we compute the sum of $Mask_{i_k}$ to obtain the number of remaining tokens after pruning at layer $i_k$. The penalty term can be expressed as:
\setlength{\abovedisplayskip}{10pt}
\setlength{\belowdisplayskip}{10pt}
\begin{align} 
\label{eq:8}
\mathcal{L}_{\text{ATP}} =  \sum^{I}_{i_k} \frac{Sum(Mask_{i_k})}{576} * i_k,
\end{align}
To constrain the average token count with a target value, we compute the average token count $\overline{N}$ across all layers within a batch and 
minimize the difference between $\overline{N}$ and the target token value $N_{\text{target}}$ using:
\setlength{\abovedisplayskip}{10pt}
\setlength{\belowdisplayskip}{10pt}
\begin{align} 
\label{eq:8_1}
\mathcal{L}_{\text{target}} = \mid\mid \overline{N} - N_{\text{target}} \mid\mid,
\end{align}
We adopt a supervised fine-tuning (SFT) setting for visual language models as our training paradigm.
The full training objective for ATP module is:
\begin{align} 
\label{eq:9}
\mathcal{L} = \mathcal{L}_{\text{ntp}} + \mathcal{L}_{\text{ATP}} * \lambda_{\text{ATP}} + \mathcal{L}_{\text{target}} * \lambda_{\text{target}},
\end{align}
where $\lambda_{\text{ATP}}$ and $\lambda_{\text{target}}$ are scaling coefficients that control the impact of computational budget constraints on training.

\subsection{Efficiency Analysis}
\label{sec:effi}
At inference time, tokens with values below the threshold are directly discarded, which reduces the computational overhead of the model during inference.
We only consider the computation of multi-head attention and feed-forward network module of LLM in the FLOPs estimation.
The theoretical FLOPs of each LLM layer with unpruned tokens can be calculated as:
\begin{align} 
\label{eq:10}
\text{FLOPs}^{L} = 4Ld^2 + 2L^2d + 2Ldm ,
\end{align}
where $L$ represents the total number of input tokens to the first layer of the LLM without pruning, $d$ denotes the hidden state size, and $m$ is the intermediate size of FFN.
Example as the ATP module is inserted after the $i_k$-th layer, $i_k \in I$.
Let $L_{i_k}^p$ represent the preserved token length after pruning, $i_N$ denote the final layer index in the LLM decoder.
We compute the FLOPs reduction across the entire model as:
\setlength{\abovedisplayskip}{4pt}
\setlength{\belowdisplayskip}{4pt}
\begin{align} 
\label{eq:11}
\sum^{\{I, i_N\}}_{i_k} (i_{k+1} - i_{k}) * (\text{FLOPs}^{L} - \text{FLOPs}^{L_{i_k}^p}) .
\end{align}
Additionally, our ATP-LLaVA introduces a slight additional computational overhead for the ATP module, which is further elaborated in the supplementary material.

%% file: sec/3_experiment.tex
\renewcommand{\arraystretch}{1}  
\begin{table*}[t]
  \centering
  \begin{tabularx}{1\textwidth}{l|>{\centering\arraybackslash}X|*{7}{>{\centering\arraybackslash}X}|>{\centering\arraybackslash}X}
    \toprule[1pt]
    Method  & Token & \textbf{GQA} & \textbf{MMB} & \textbf{MME} & \textbf{POPE} & \textbf{SEED} & \textbf{SQA$^{I}$} & \textbf{VQA$^{v2}$} & \textbf{Avg.} \\
    \midrule[1pt]
    \multicolumn{10}{c}{{\emph{Upper Bound Model}}} \\
    \midrule[1pt]
    
    \multirow{2}{*}{LLaVA-1.5~\cite{liu2023improvedllava}} & \multirow{2}{*}{576} & 62.0 & 64.3 & 1510.7 & 85.8 & 58.6 & 71.6 & 78.5 & -  \\ 
      && 100{\footnotesize \%} & 100{\footnotesize \%} & 100{\footnotesize \%} & 100{\footnotesize \%} & 100{\footnotesize \%} & 100{\footnotesize \%} & 100{\footnotesize \%} & 100{\footnotesize \%}\\
     
    \midrule[1pt]
    \multicolumn{10}{c}{{\emph{Methods with Pre-defined Pruning Ratio}}} \\
    \midrule[1pt]
    \multirow{2}{*}{{PruMerge+~\cite{shang2024LLaVA-PruMerge}}} & \multirow{2}{*}{144}   & {-} & \underline{64.9} & \underline{1462.4} & \underline{84.0} & {-} & \underline{68.3} & {\textbf{76.8}} & - \\
    & & {-} & {100.9{\footnotesize \%}} & {96.8{\footnotesize \%}} & {97.9{\footnotesize \%}} & {-} & {95.4{\footnotesize \%}} & {97.8{\footnotesize \%}} & {97.8{\footnotesize \%}} \\
    \midrule
    
    \multirow{4}{*}{{FastV~\cite{chen2024image}}} & \multirow{2}{*}{192}   & {52.7} & {61.2} & {1312.4} & {64.8} & {50.8} & {65.4} & {67.1} & - \\
    & & {83.8{\footnotesize \%}} & {95.2{\footnotesize \%}} & {86.9{\footnotesize \%}} & {75.5{\footnotesize \%}} & {86.7{\footnotesize \%}} & {91.3{\footnotesize \%}} & {85.5{\footnotesize \%}} & {86.4{\footnotesize \%}} \\
      \cmidrule{2-10}
     & \multirow{2}{*}{128}   & {49.6} & {56.1} & {1187.9} & {59.6} & {48.1} & {59.7} & {61.8} & - \\
    & & {80.0{\footnotesize \%}} & {87.2{\footnotesize \%}} & {78.6{\footnotesize \%}} & {69.5{\footnotesize \%}} & {82.1{\footnotesize \%}} & {83.4{\footnotesize \%}} & {78.7{\footnotesize \%}} & {79.9{\footnotesize \%}} \\
    \midrule
    
    \multirow{4}{*}{{SparseVLM~\cite{zhang2024sparsevlm}}} & \multirow{2}{*}{192}   & \underline{57.6} & {62.5} & {1382.8} & {83.6} & {53.0} & {67.2} & {75.6} & - \\
    & & {92.9{\footnotesize \%}} & {97.2{\footnotesize \%}} & {91.5{\footnotesize \%}} & {97.4{\footnotesize \%}} & {90.4{\footnotesize \%}} & {93.9{\footnotesize \%}} & {96.3{\footnotesize \%}} & {94.2{\footnotesize \%}} \\
      \cmidrule{2-10}
    
     & \multirow{2}{*}{128}   & {56.0} & {60.0} & {1296.7} & {80.5} & {50.2} & {65.5} & {73.8} & - \\
    & & {90.3{\footnotesize \%}} & {93.3{\footnotesize \%}} & {85.8{\footnotesize \%}} & {96.3{\footnotesize \%}} & {85.7{\footnotesize \%}} & {91.5{\footnotesize \%}} & {94.0{\footnotesize \%}} & {91.0{\footnotesize \%}} \\
    \midrule[1pt]
    \multicolumn{10}{c}{{\emph{Methods with Adaptive Pruning Ratio}}} \\
    \midrule[1pt]
    
    \multirow{4}{*}{\textbf{ATP-LLaVA}} & \multirow{2}{*}{144\textsuperscript{*}}  & {\textbf{59.5}} & {\textbf{66.0}} & {\textbf{1473.9}} & {\textbf{84.2}} & {\textbf{57.3}} & {\textbf{69.1}} & {\underline{76.4}} & - \\
    & & {96.0{\footnotesize \%}} & {102.6{\footnotesize \%}} & {97.6{\footnotesize \%}} & {98.1{\footnotesize \%}} & {97.8{\footnotesize \%}} & {96.5{\footnotesize \%}} & {97.3{\footnotesize \%}} & \textbf{98.1{\footnotesize \%}} \\
    
      \cmidrule{2-10}
     & \multirow{2}{*}{88\textsuperscript{*}}  & {56.8} & {64.7} & {1401.5} & {82.6} & \underline{55.7} & {67.2} & {73.3} & - \\
    & & {91.6{\footnotesize \%}} & {100.6{\footnotesize \%}} & {92.8{\footnotesize \%}} & {96.3{\footnotesize \%}} & {95.1{\footnotesize \%}} & {93.9{\footnotesize \%}} & {91.8{\footnotesize \%}} & \textbf{94.6{\footnotesize \%}} \\
    
    \bottomrule[1pt]
  \end{tabularx}
  \caption{
  Comparison with previous approaches on vision token pruning within LLM decoder layers using common visual understanding benchmarks.
  (Token) indicates the average token count for all layers in language model.
  Token count with (*) means the retained token count is not pre-defined and adaptively determined by learnable ATP module. 
  We calculate the average token count during inference across all benchmarks.
  The percentage represents the compression retention rates to Upper Bound model.
}
\label{tab:t_1}
\end{table*}

\section{Experiments}
\subsection{Implementation Details} \label{sec:implementation details}
Regarding the training strategy and data, ATP-LLaVA follows a standard vision instruction tuning stage.
LLaVA-1.5~\cite{liu2023improvedllava} is chosen as the base model for ATP-LLaVA.
Specifically, we employ the pre-trained CLIP-ViT-L~\cite{Radford2021LearningTV} as our visual encoder followed by a linear projector to align text and vision modality.
We directly utilize pretrained projectors from LLaVA-1.5, which were trained on the filtered CC3M dataset~\cite{sharma2018conceptual} with the fixed language model and vision encoder.
For pre-trained large language models, we utilize Vicuna-7B-1.5~\cite{vicuna2023}. 
For training data, we use 665k visual instruction following data~\cite{liu2023improvedllava} to tune our model.
We train ATP-LLaVA using lr of 2e-5 for LLM and 1e-4 for the ATP module for 1 epoch.
All other hyperparameters and settings are identical to those used in LLaVA-1.5.
As for the scaling coefficients, $\lambda_{sample}$ is set as 3, $\lambda_{\text{target}}$ is set as 0.2, and $\lambda_{\text{ATP}}$ is set within the range of 0.01 to 0.1.
Varying the value of $\lambda_{\text{ATP}}$ during model training has an impact on the final average FLOPs of the resulting model.

\subsection{Datasets} 
We conduct experiments on several common visual understanding benchmarks for vision token pruning in this work.
Specifically, we report results on GQA~\cite{hudson2018gqa}, MMB (MMBench)~\cite{MMBench}, MME~\cite{fu2023mme}, POPE~\cite{Li2023pope}, SEED-Bench~\cite{li2023seed}, SQA$^I$ (Image-based setting in ScienceQA)~\cite{lu2022sqa} and VQA$^{v2}$ (VQA V2)~\cite{balanced_vqa_v2}.
We can assess the impact of visual information loss during the pruning process by comparing the model's performance on these visual understanding benchmarks before and after pruning.
We follow the evaluation details outlined in~\cite{liu2023llava} to assess the model's performance on these visual understanding benchmarks.

\subsection{Results}

\renewcommand{\arraystretch}{1} 
\begin{table*}[t]
  \centering
  \small 
  \begin{tabular}{l|c|c|c|c|cccc}
    \toprule[1pt]
    Pruning & Avg. & Pruning & Pruning & Avg.  & \multirow{2}{*}{\textbf{MMB}} & \multirow{2}{*}{\textbf{GQA}} & \multirow{2}{*}{\textbf{VQA$^{v2}$}} & \multirow{2}{*}{\textbf{SEED}}  \\
     Strategy & Tokens & Ratio & Layer Indexes & Retained Tokens &  &  &  &   \\
    \midrule[1pt]
    Upper Bound & 576 & - & - & - & 64.3 & 62.0 & 78.5 & 58.6     \\
    \midrule
    \multirow{3}{*}{Pre-defined Ratio} & 144 & $3/4$ & [1]        & [130]         & 59.6 & 56.2 & 71.6 & 53.1     \\
                                      & 144 & $2/3$ & [4, 14, 24] & [162, 54, 18] & 62.7 & 58.0 & \underline{74.6} & \underline{55.7}     \\
                                      & 144 & $1/2$ & [4, 14, 24] & [136, 68, 34] & 63.2 & \underline{58.2} & {73.9} & 55.3     \\
    \midrule
    \multirow{2}{*}{\textbf{ATP-LLaVA}}  & 144\textsuperscript{*} & \multirow{2}{*}{-} & [4, 14, 24]  & [{126\textsuperscript{*}}, {88\textsuperscript{*}}, {20\textsuperscript{*}}] & \textbf{66.0} & \textbf{59.5} & \textbf{76.4} &\textbf{57.3}     \\
    
     & 88\textsuperscript{*} &  & [1, 13, 25]  & [{98\textsuperscript{*}}, {79\textsuperscript{*}}, {16\textsuperscript{*}}] & \underline{64.7} & 56.8 & 73.3 & \underline{55.7}     \\
    \bottomrule[1pt]
  \end{tabular}
  \caption{Comparison with pre-defined pruning strategy under same traing setting using common benchmarks.
  (Pruning Layer Indexes) specify the LLM layer indexes at which vision tokens are pruned, starting from 0 and occurring prior to input.}
\vspace{-4pt} 
\label{tab:t_2}
\end{table*}

\renewcommand{\arraystretch}{1} 
\begin{table}[t]
  \centering
  \small
\resizebox{1\columnwidth}{!}{
  \begin{tabular}{l|c|cccc}
    \toprule[1pt]
    LM & Token & {\textbf{MMB}} & {\textbf{GQA}} & {\textbf{VQA$^{v2}$}} & {\textbf{SEED}}  \\
    \midrule[1pt]
    \textit{Frozen} & 144\textsuperscript{*} & 63.1 & 57.9 & 75.2 & 54.9     \\
    
    \textit{Trainable} & 144\textsuperscript{*} & {66.0} & {59.5} & {76.4} &{57.3}     \\
    \bottomrule[1pt]
  \end{tabular}}
  \caption{Ablation study on training strategy for ATP-LLaVA using common visual understanding benchmarks. 
  (\textit{Frozen}) means only the ATP module is trainable while freezing the language model.}
\label{tab:t_3}
\end{table}

\renewcommand{\arraystretch}{1} 
\begin{table}[t]
  \centering
  \small 
\resizebox{1\columnwidth}{!}{
  \begin{tabular}{c|c|c|cccc}
    \toprule[1pt]
     RP & SP & PP & {\textbf{MMB}} & {\textbf{GQA}} & {\textbf{VQA$^{v2}$}} & {\textbf{SEED}}  \\
    \midrule[1pt]
    \checkmark & \checkmark & \checkmark  & \textbf{66.0} & \textbf{59.5} & \textbf{76.4} & \textbf{57.3}     \\
    \checkmark &  &                       & \underline{65.3} & 58.3 & \underline{75.1} & \underline{56.5}     \\
     & \checkmark &                       & 64.3 & 57.1 & 74.2 & 55.1     \\
     & \checkmark & \checkmark            & 65.2 & \underline{58.6} & {74.9} & {56.3}     \\
    \bottomrule[1pt]
  \end{tabular}}
  \caption{Ablation study on SAP approach for ATP-LLaVA using common visual understanding benchmarks. 
  (RP) indicates using redundant pruning strategy, (SP) indicates using spatial pruning strategy, and (PP) indicates using pruning positional embedding. }
\label{tab:t_4}
\end{table}

\renewcommand{\arraystretch}{1}  
\begin{table}[t]
  \centering
  \small 
\resizebox{1\columnwidth}{!}{%
  \begin{tabular}{c|c|cccc}
    \toprule[1pt]
     $S^{self}$ & $S^{cross}$ & {\textbf{MMB}} & {\textbf{GQA}} & {\textbf{VQA$^{v2}$}} & {\textbf{SEED}}  \\
    \midrule[1pt]
     \checkmark & \checkmark    & \underline{66.0} & \textbf{59.5} & \textbf{76.4} & \textbf{57.3}     \\
    \checkmark &                & \textbf{66.1} & {58.7} & {75.8} & \underline{56.9}     \\
     & \checkmark               & {65.4} & \underline{59.2} & \underline{76.1} & {56.3}     \\
    \bottomrule[1pt]
  \end{tabular}}
  \caption{Ablation study on redundant pruning score for ATP-LLaVA using common visual understanding benchmarks. 
  ($S^{self}$) indicates the self-modality importance score in~\cref{eq:3}, and ($S^{cross}$) indicates the cross-modality importance score in~\cref{eq:4}. }
\vspace{-4pt} 
\label{tab:t_5}
\end{table}

\renewcommand{\arraystretch}{1}  
\begin{table*}[t]
  \centering
  \small 
  \begin{tabular}{l|c|c|c@{\hspace{0.5em}}c|c@{\hspace{0.5em}}c|c@{\hspace{0.5em}}c}
    \toprule[1pt]
    \multirow{2}{*}{Method} & Avg. & \multirow{2}{*}{Accuracy} &  Storage & \multirow{2}{*}{\(\Delta\)} & CUDA  & \multirow{2}{*}{\(\Delta\)} & FLOPs & \multirow{2}{*}{\(\Delta\)} \\
     & Token & & Memory (MB) &  & Time (ms) $\downarrow$ &  & (T) $\downarrow$ &  \\
    \midrule[1pt]
    Upper Bound                & 576 & 100{\footnotesize \%}  & 302.4 & -      & 432.7 & -                      & 9.6 & -    \\
    FastV~\cite{chen2024image} & 144 & 81.7{\footnotesize \%}   & 75.6 & 75\% & 259.1 & 40.1{\footnotesize \%} & 2.0 & 79.2{\footnotesize \%}     \\
    \midrule
    \multirow{2}{*}{\textbf{ATP-LLaVA}} & 144 & \textbf{98.1{\footnotesize \%}}   & 75.6 & 75\% & 266.4 & 38.4{\footnotesize \%} & 2.1 & 78.1{\footnotesize \%}     \\
     & 88 & \textbf{94.6{\footnotesize \%}}   & 46.2 & 84.7\% & 226.8 & 47.6{\footnotesize \%} & 1.5 & 84.4{\footnotesize \%}     \\
    \bottomrule[1pt]
  \end{tabular}
  \caption{Efficiency analysis of ATP-LLaVA including cache storage memory, CUDA times and the FLOPs. \(\Delta\) denotes the reduction ratio.}
\vspace{-4pt} 
\label{tab:t_7}
\end{table*}

\noindent\textbf{Main Results.}
We report results of our ATP-LLaVA on various common visual understanding benchmarks to presents the vision token pruning performance.
Besides, to rigorously quantify the performance loss of ATP-LLaVA during token pruning, we also report the compression retention rates to the Upper Bound model (\emph{i.e.}, LLaVA-1.5~\cite{liu2023improvedllava} in this paper).
We compare our method with previous token pruning methods.
For fair comparisons, we constrain ATP-LLaVA with a limited budget and average token count.
In particular, we adjust the scaling coefficients and the target token number in~\cref{eq:8_1} and \cref{eq:9}.
The token count is an average value across all decoder layers (\emph{i.e.}, 32 layers in LLaMA~\cite{touvron2023llama}) on a uniform sampled set across all reported benchmarks during inference.
As shown in \cref{tab:t_1}, it can be observed that our method preserves the original vision understanding capability to a large extent.
Notably, we achieved an average compression retention rate of 98.1\% and 94.6\% across seven widely used benchmarks, when pruning from 576 to 144 and 88 tokens, respectively.
Especially on MMBench and SEED-Bench, our method achieves and even surpasses the performance of the Upper Bound model.
This demonstrates that ATP-LLaVA can substantially enhance inference efficiency with only a negligible performance degradation.

\noindent\textbf{Pruning Strategy.}
To evaluate the adaptability of Adaptive Token Pruning (ATP) module, we conduct experiments using pre-defined pruning ratio strategy.
Specifically, we replace the ATP module with a pre-defined pruning ratio strategy that prunes visual tokens at a fixed ratio across all layers, while maintaining the other modules.
The pruned tokens will be directly discarded, eliminating the need of differentiable design introduced in~\cref{sec:learnable}.
The pre-defined ratio models are trained under the same setting and dataset as ATP-LLaVA.
We report the performance of the model on common visual understanding benchmarks at various pruning rates ($3/4$, $2/3$, and $1/2$) and compare it to the ATP strategy with the same average number of pruned tokens.
As shown in~\cref{tab:t_2}, ATP-LLaVA consistently outperforms models trained with fixed pruning rates across all four benchmarks. 
Especially, ATP-LLaVA attains noticeable improvements compared with the best fixed pruning ratio model, with large margins of 2.8, 1.6 and 1.6 absolute points on MMBench, VQA$^{v2}$ and SEED. 
These results illustrate that tailoring adaptive pruning strategies to specific instances and layers can help mitigate performance degradation when pruning tokens within a constrained budget.

\begin{figure*}[t]
\centering
\includegraphics[width=1\linewidth]{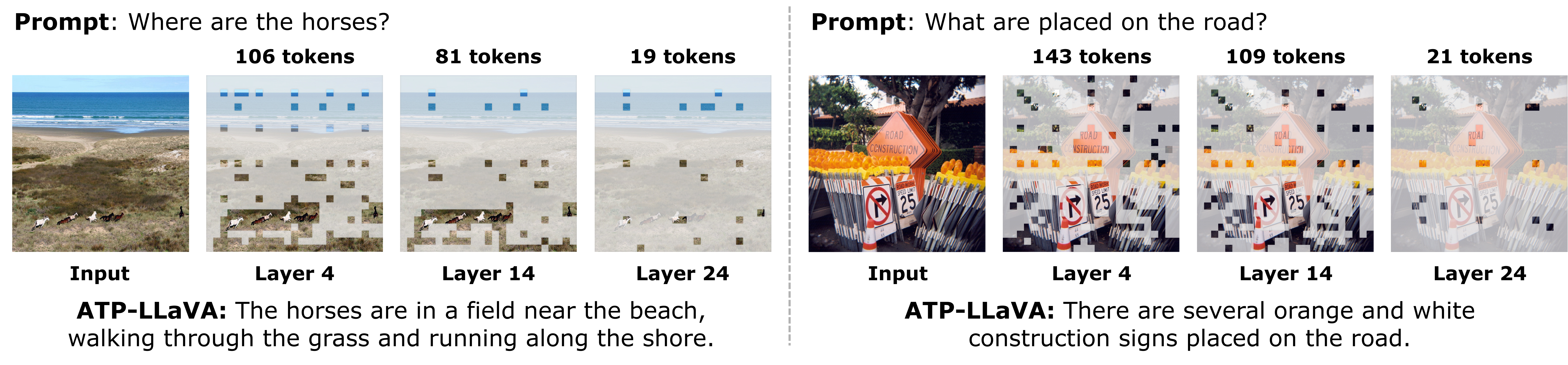}
\vspace{-12pt}
\caption{
Visualized vision token pruning results of ATP-LLaVA. 
White tokens represents the pruned tokens.
The uniform sampled tokens are pruned by spatial pruning threshold, while the sparse tokens are pruned by redundant pruning threshold.
Zoom in to have a better view.
}
\vspace{-4pt}
\label{fig:i3}
\end{figure*}

\noindent\textbf{Training Strategy.}
We further investigate the influence of training strategies on the performance of ATP-LLaVA. 
In our main experiments, we adopt the training setting of visual instruction tuning, where both the ATP module and language model are trainable. 
To explore more efficient training paradigms, we conduct additional experiments by freezing the language model and solely training the lightweight ATP module. 
We omit pruning positional embedding from these comparative experiments, as it is incompatible with the frozen language model training paradigm.
As evident in~\cref{tab:t_3}, ATP-LLaVA is capable of learning adaptive token pruning strategies and mitigating the degradation in capability caused by token pruning under the training paradigm with a frozen language model. 
Concurrently, fine-tuning the language model enables the LLM to better adapt to visual understanding with a limited number of tokens.
More training details can be found in the supplementary materials.

\noindent\textbf{Token Pruning Technique.}
We conduct several ablations to evaluate the effectiveness of the key components in ATP-LLaVA.
Firstly, we remove the redundant pruning strategy (RP), spatial pruning strategy (SP) and pruning positional embedding (PP), respectively.
The average token count in these experiments is adjusted to be identical, \emph{i.e.}, 144.
\cref{tab:t_3} shows that removing redundant pruning and spatial pruning leads to an average drop of 1.05 and 1.0 absolute points across four common benchmarks, respectively.
Furthermore, in comparison to employing spatial sampling pruning in isolation, the synergistic application of spatial pruning and pruning positional encoding yields more pronounced performance improvements.
Specifically, the integration of pruning positional encoding results in an improvement of 0.9 absolute points when spatial pruning strategy is applied.
These results demonstrate the benefit of exploiting the pruning technique from both token redundancy and spatial modeling perspective.

\noindent\textbf{Pruning Importance Score.}
In this ablation study, we separately utilize either the self-modality importance score or the cross-modality importance score for redundant token pruning. 
As shown in~\cref{tab:t_5}, relying solely on self-score and cross-score results in average absolute performance drops of 0.43 and 0.55 points, respectively, across four common benchmarks.
These results illustrate that both intra-modal and cross-modal importance scores are equally significant in quantifying the redundancy of visual tokens.

\noindent\textbf{Efficiency Analysis.}
We evaluate the inference efficiency of ATP-LLaVA through a comparative analysis with the Upper Bound model and FastV~\cite{chen2024image}. 
Our results, presented in~\cref{tab:t_7}, demonstrate that ATP-LLaVA achieves significant reductions in KV cache storage memory (75\%), CUDA time (38.4\%), and FLOPs (78.1\%) compared to the Upper Bound model, while maintaining 98.1\% performance.
In comparison to FastV, the introduction of threshold prediction linear layers in ATP module incurs a minor computational overhead (1.1\% in FLOPs, 1.7\% in CUDA time), but yields a substantial performance gain (16.4\% accuracy improvement). 
We conclude that the performance benefits outweigh the negligible increase in computational cost.
More discussions are provided in the supplementary materials.

\noindent\textbf{Visualization Results.}
In~\cref{fig:i3}, we visualize the instance-specific pruning results using the optimal ATP-LLaVA model from~\cref{tab:t_1}, which has an average token count of 144. 
The uniformly distributed tokens in the figure result from spatial token pruning, whereas the sparse tokens arise from redundancy token pruning. 
The example on the left illustrates an image of lower complexity, where the model prunes extensively in the shallow and middle layers (4-th and 14-th layers), while selectively retaining prompt-relevant tokens (e.g., ``horses'').
In contrast, the example on the right presents an image of higher complexity, where the model retains a larger number of tokens in the shallow and middle layers, particularly those related to the object's location (\emph{e.g.}, ``on the road''). 
In the 24-th layer, token preservation is minimal, as visual information has been largely consolidated into text tokens through preceding layers. 
These results demonstrate that ATP-LLaVA adaptively prunes tokens across layers based on instance-specific visual complexity and text-vision relevance. 
We provide additional visualizations in the supplementary materials.

%% file: sec/4_conclusion.tex
\section{Conclusion} \label{sec:conclusion}
In this paper, we propose ATP-LLaVA, which adaptively prunes vision tokens for large vision language models on layer and instance level.
By introducing an Adaptive Token Pruning module, our method can calculate the importance score for vision tokens and determine the pruning threshold dynamically.
Based on Spatial Augmented Pruning strategy, our method can prune redundant vision tokens and maintain spatial modeling, while minimizing information loss.
In summary, our approach offers a promising solution for flexible vision token pruning of LVLMs, making them more scalable in resource constrained-environments.